\documentclass{article}

\usepackage{arxiv}

\usepackage[utf8]{inputenc} % allow utf-8 input
\usepackage[T1]{fontenc}    % use 8-bit T1 fonts
\usepackage{hyperref}       % hyperlinks
\usepackage{url}            % simple URL typesetting
\usepackage{booktabs}       % professional-quality tables
\usepackage{amsfonts}       % blackboard math symbols
\usepackage{nicefrac}       % compact symbols for 1/2, etc.
\usepackage{microtype}      % microtypography
\usepackage{lipsum}		% Can be removed after putting your text content
\usepackage{graphicx}
\usepackage[square,numbers]{natbib}
\usepackage{doi}

\usepackage{lineno,hyperref}

\usepackage{subfig}
\usepackage{comment}
\usepackage{graphicx}
\graphicspath{{images/}}
\usepackage{booktabs,array}
\usepackage{amsmath}
\usepackage{algorithm}
\usepackage{algpseudocode}
\usepackage{multirow}
\usepackage{multicol}
\usepackage{tabularx}
\title{Forecasting of COVID-19 Cases, Using an Evolutionary Neural Architecture Search Approach}

\author{ \href{https://orcid.org/0000-0001-8718-9755}{\includegraphics[scale=0.06]{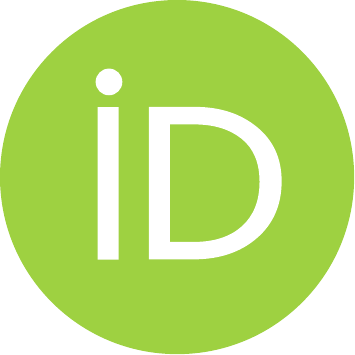}\hspace{1mm}Mahdi Rahbar}\\
	Department of Computer Science\\
	Saint Louis University\\
	St. Louis, MO 63103 \\
	\texttt{mahdi.rahbar@slu.edu} \\
	\And
	\href{https://orcid.org/0000-0001-7806-5152}{\includegraphics[scale=0.06]{orcid.pdf}\hspace{1mm}Samaneh Yazdani, Ph.D.} \\
	Department of Computer Engineering\\
	North Tehran Branch, Islamic Azad University\\
	Tehran, Iran \\
	\texttt{s\_yazdani@iau-tnb.ac.edu} \\
}

\renewcommand{\shorttitle}{\textit{arXiv} Forecasting of COVID-19 Cases Using Evolutionary NAS}  % \textit{arXiv}

\hypersetup{
	pdftitle={Forecasting of COVID-19 Cases, Using an Evolutionary Neural Architecture Search Approach},
	pdfsubject={cs.AI , cs.NE, cs.DS, cs.LG,q-bio.PE,q-bio.QM},
	pdfauthor={Mahdi Rahbar, Samaneh Yazdani},
	pdfkeywords={Neural Architecture Search, Feature Augmentation,  Forecasting COVID-19 Cases, Deep Learning},
}

\begin{document}
\maketitle

\begin{abstract}
	In late 2019, COVID-19, a severe respiratory disease, emerged, and since then, the world has been facing a deadly pandemic caused by it. This ongoing pandemic has had a significant effect on different aspects of societies. The uncertainty around the number of daily cases made it difficult for decision-makers to control the outbreak. Deep Learning models have proved that they can come in handy in many real-world problems such as healthcare ones. However, they require a lot of data to learn the features properly and output an acceptable solution. Since COVID-19 has been a lately emerged disease, there was not much data available, especially in the first stage of the pandemic, and this shortage of data makes it challenging to design an optimized model. To overcome these problems, we first introduce a new dataset with augmented features and then forecast COVID-19 cases with a new approach, using an evolutionary neural architecture search with Binary Bat Algorithm (BBA) to generate an optimized deep recurrent network. Finally, to show our approach's effectiveness, we conducted a comparative study on Iran's COVID-19 daily cases. The results prove our approach's capability to generate an accurate deep architecture to forecast the pandemic cases, even in the early stages with limited data.
\end{abstract}

% keywords can be removed
\keywords{Neural Architecture Search \and Feature Augmentation \and Forecasting COVID-19 Cases \and Deep Learning}

	\section{Introduction}

The recent worldwide health challenge caused by severe acute respiratory syndrome coronavirus 2 (SARS-CoV-2) has caused a lot of fear and uncertainty for humanity. SARS-CoV-2 is a genetic variant of coronavirus that causes coronavirus disease 2019 (COVID-19). The crisis that governments started to face in the early stage of this phenomenon was controlling the pandemic and Covid-19 outbreak alongside maintaining economic balance and other aspects of governmental matters. One of the essentials to assist decision-makers in developing better solutions has been analyzing the pandemic growth and forecasting Covid-19 cases. 

Accurate prediction of Covid-19 daily cases can assist governments with macro-decisions and controlling the pandemic better. Meanwhile, artificial intelligence techniques have proven that they are capable and accurate in finding patterns from indistinctive and complicated data features in different phenomena, such as pandemic epidemiological studies. Since the emergence of SARS-CoV-2, researchers have applied various techniques to study different aspects of the current pandemic, such as predicting COVID-19 cases growth rate. 

Although multiple techniques such as \cite{Pathan2020,Arora2020} have utilized a variant of Recurrent Neural Networks (RNN) to predict daily cases, their proposed models have two shortcomings. Firstly, many studies \cite{Lee2020,Hawas2020} have chosen the framing range by assuming a fixed specific number. However, RNN and LSTM require a proven best time step for framing the sequence data that guarantees sufficient distinctive features, and on the other hand, it prevents adding too much data to mislead the model. In other words, by controlling the amount of the sequence information, we try to provide the model with the most informative data sequence for training without feeding it extra data that can cause noise. 
Secondly, they utilized customized architectures obtained by trial and error, which might still not be the best topology chosen. As a result of these two main factors, there will be so much inaccuracy in prediction.

In this paper, we have taken a deep neuroevolutionary approach, using the Binary Bat algorithm to optimize the hyperparameters of a recurrent neural network with Long Short-Term Memory (LSTM) layers to predict daily cases. Hyperparameters optimization is an NP-hard problem as the optimal solution cannot be guaranteed to be obtained unless by performing an exhaustive search in the feasible region. Therefore, we have chosen the BBA algorithm as a well-known metaheuristic technique for exploring the best set of hyperparameters in the search space.
This approach helps us obtain the optimum time-sequence as well as the most optimized architecture for our deep learning framework. We also introduce a new feature augmentation version of the latest available public COVID-19 dataset provided by the European Center for Disease Prevention and Control. It will be shown that the model's accuracy is increased with the help of the new features and can simulate the regional pandemic behavior more precisely. To validate the framework and the final model, we have conducted various experiments that, in all cases, show the effectiveness of our approach.

In the following sections, we first investigated the related works and briefly talked about the background. In section 3, our proposed model is explained in detail, and we discussed why this approach had been taken for forecasting COVID-19 cases. In section 4, experimental results are presented and investigated in detail, and finally, in section 5, we discussed the conclusion and possible future works.

\section{Related works}\label{relatedworkssection}
There are many studies on applications of artificial intelligence for the Covid-19 pandemic \cite{Lalmuanawma2020, Ke2020, Tuli2020}. One of the main topics among these studies is predicting new cases to help health managers plan and develop appropriate strategies to deal with the Covid-19. Here we study some of them:

ArunKumar et al. \cite{ArunKumar2021} predicted the future trends of the cumulative fatalities of the top 10 countries in the range of 60 days, using RNN along with Gated Recurrent Units (GRUs) and Long Short-Term Memory (LSTM).

\cite{Gautam2021} utilized transfer learning in LSTM networks to forecast COVID-19 cases using the early COVID infected countries such as Italy and used the learned model to predict cases in other countries. The results of the model on multiple countries showed the effectiveness of this approach. 

Shastri et al. \cite{Shastri2021} proposed a nested ensemble model using LSTM to enhance the accuracy of predicting daily cases of India. 

Abbasimehr et al. \cite{Abbasimehr2021} studied three different hybrid deep models, namely multi-head attention, LSTM, and CNN, optimized with a Bayesian algorithm to forecast COVID-19 cases. The results showed the superiority of their proposed model among the studied benchmark models.

\cite{Chandra2021} used LSTM, bidirectional LSTM and encoder-decoder LSTM models for multi-step forecasting of COVID-19 two-month ahead cases in India. They claimed that the deep models are promising in terms of finding the long-term prediction of cases.

Salgotra et al. \cite{Salgotra2020} utilized gene expression programming (GEP) to present a model for predicting confirmed cases (CC) and death cases (DC) in the fifteen most affected countries of the world. Two GEP models were introduced for CC and DC for all 15 countries. The results were shown that GEP provides better results than neural network models when the total experimental data is limited.

To estimate the possible spread of the Coronavirus 2 (SARS-CoV-2) in three Indian cities, a new GEP based model was presented in \cite{Salgotra2020a}. The proposed model is utilized to predict the total number of cases based on CC, DC, and the other three parameters.

% Deep learning methods for forecasting COVID-19 time-series data, Zeroual et al. \cite{Zeroual2020}eep learning methods to predicate the number of new cases and recovered ones. % % The results of collected data from six countries confirmed the performance of deep learning to the promising forecasting COVID-19 cases. 

In \cite{Chimmula2020}, LSTM was used to predict the trends and possible stopping time of the current COVID-19 outbreaks in Canada. Since COVID-19 is a time series dataset, sequential networks are useful to extract a pattern from it. In \cite{Chimmula2020}, the internal connections of LSTM were established to improve its performance. The results show that the ending point of the COVID-19 outbreak was predicted in June 2020 in Canada. 

To enhance the public health management in dealing with the COVID-19 in two high daily incidences of new cases and deaths, \cite{daSilva2020} applied some machine learning algorithms such as quantile random forest and support vector regression to forecast one, three, and six-days-ahead the Covid-19 cumulative cases.

%	\begin{tabular}{l|l|l|l}  % \begin{tabular}{l*{6}{c}r}
%		\toprule
%	    Author   &  Method  &   Target  &  Results \\
%		\hline
%		\midrule
%		Salgotra et al. [1] & Genetic Evolutionary Programming (GEP) & \parbox{5cm}{Betty Botter Thought: \\ If I Put This Bitter Butter in My Batter it Will Make My Batter Bitter} & \parbox{5cm}{Betty Botter Thought: \\ If I Put This Bitter Butter in My Batter it Will Make My Batter Bitter}  \\
%		Salgotra et al. [1] & Genetic Evolutionary Programming (GEP) & Predicting confirmed cases and death cases & Predicting confirmed cases and death cases for next 10 days \\
%		Salgotra et al. [1] & Genetic Evolutionary Programming (GEP) & Predicting confirmed cases and death cases & Predicting confirmed cases and death cases for next 10 days  \\
%		\bottomrule
%	\end{tabular}

\section{Background}\label{idbackgroundsection}

\subsection{Binary Bat Algorithm}
Bat algorithm has been inspired by simplification and simulation of the echolocation capability of bats in 2010 \cite{Yang2010}. Similar to other population-based algorithms, BA starts with randomly generated individuals. In BA, each bat represents an individual, which is a solution in the search space. Each bat can be represented by a group of vectors: frequency, velocity, and position. For the \textit{i}th bat these vectors are updated according to equation \eqref{eq:Eq.1}, \eqref{eq:Eq.2} and \eqref{eq:Eq.3} respectively.

\begin{equation} \label{eq:Eq.1}
	f_i=f_min+\beta(f_max-f_min)
\end{equation}
\begin{equation} \label{eq:Eq.2}
	v_i\ (t)=v_i\ (t-1)+f_i\ (x_i\ (t)-x^\ast)
\end{equation}
\begin{equation} \label{eq:Eq.3}
	x_i\ (t)=x_i\ (t-1)+v_i\ (t)
\end{equation}

Where $f_\textit{i}$ is the frequency of $i$th bat, $f_{max}$ and $f_{min}$ show the maximum and minimum value of frequency respectively. $\beta$ is random number in the interval [0,1]. $v_i$ and $x_i$ indicate the velocity and position of the $i$th bat. $x^\ast$ shows the best position by the entire population so far. Algorithm~\ref{alg:BA} shows the pseudo-code of the basic BA.

\begin{algorithm}
	\caption{Pseudo-code of BA \cite{Yang2010}}\label{alg:BA}
	\begin{algorithmic}[1]  % \scriptsize   \footnotesize
		{\footnotesize
			\Procedure{BA}{}
			\State 	Initialize the position, velocity and frequency of bats $(x_i.v_i.f_i\ \ i=1\cdots n)$.
			\Repeat:
			\State Update frequencies, velocities and positions of bats using Eqs.\eqref{eq:Eq.1} to \eqref{eq:Eq.3}.
			
			\State {\bf if}  $rand>r_i$  	
			\State \hspace{10pt}Select a solution among the best solutions 
			\State \hspace{10pt}Generate a local solution around the best solution
			\State end
			
			\State {\bf if} $rand<A_i$ and $f(x_i)<f(x^\ast)$
			\State \hspace{10pt}Accept the new solutions
			\State \hspace{10pt}Modify the value of $r_i$ and $A_i$
			\State end
			
			\State Rank the bats and update the $x^\ast$ 
			
			%		\Repeat
			%		\State Select the largest principle component.% from the remaining components
			%		\State Update the selected variance fractions.% for selected components
			\Until{the stop criterion is satisfied}
			\EndProcedure
		}
	\end{algorithmic}
\end{algorithm}
\vspace{-0.0em}

where \textit{n} is the number of bats (the population size) and $rand$ is a uniformly distributed random real number in the range [0,1]. \textit{r} is pulse emission rate and increase over the course of iteration by the following equation:

\begin{equation} \label{eq:Eq.4}
	r_i(t+1)=r_i(0)(1-e^{\gamma t})
\end{equation}

Where $\gamma$ is constant and $r_i(0)$ shows the initial pulse emission rate of \textit{i}th bat. BA utilizes a local search (lines 5-8) to create a solution near the obtained ones.

\begin{equation} \label{eq:Eq.5}
	x_{new}=x_{old}+\varepsilon\bar{A}(t)
\end{equation}

In Eq.\eqref{eq:Eq.5}, $x_{old}$ is one of the current best solutions which is selected by some selection mechanism. $\varepsilon$ is a random number in the interval [-1,1], and A is the average loudness of all bats, which is calculated as follows:

\begin{equation} \label{eq:Eq.6}
	A_i(t+1)=\alpha A_i(t)
\end{equation}

Based on Eq.\eqref{eq:Eq.6} loudness $A_i$ is decreased as the iteration processed. $\alpha$ is similar to the cooling factor in simulated annealing \cite{Yang2010}. 
The basic BA was developed for solving continuous problems \cite{Yang2010}. A binary version of BA (BBA) was developed in \cite{Mirjalili2014}. BBA employs a v-shaped transfer function to transfer all real-valued velocities to the range of [0,1] as follows:

\begin{equation} \label{eq:Eq.7}
	V\left(v_{ij}\left(t\right)\right)=\left|\frac{2}{\pi}\arctan(\frac{\pi}{2}v_{ij}\left(t\right))\right|
\end{equation}

where $v_{ij}\left(t\right)$ show the \textit{j}th element of vector $v_i$ at iteration \textit{t}. In BBA, the rule for updating bat's position is redefined as:

\begin{equation} \label{eq:Eq.8}
	x_{ij}\left(t+1\right)=\left\{\begin{matrix}{(x_{ij}\left(t\right))}^{-1}&\ \ \ \ if\ rand<V\left(v_{ij}\left(t+1\right)\right)\\x_{ij}\left(t\right)\ \ \ \ \ \ &rand\geq V\left(v_{ij}\left(t+1\right)\right)\\\end{matrix}\right.
\end{equation}

\subsection{Deep Recurrent Networks}
\begin{figure}[h]
	%	\vspace{-1.0em}
	\centering
	\includegraphics[width=0.46\textwidth]{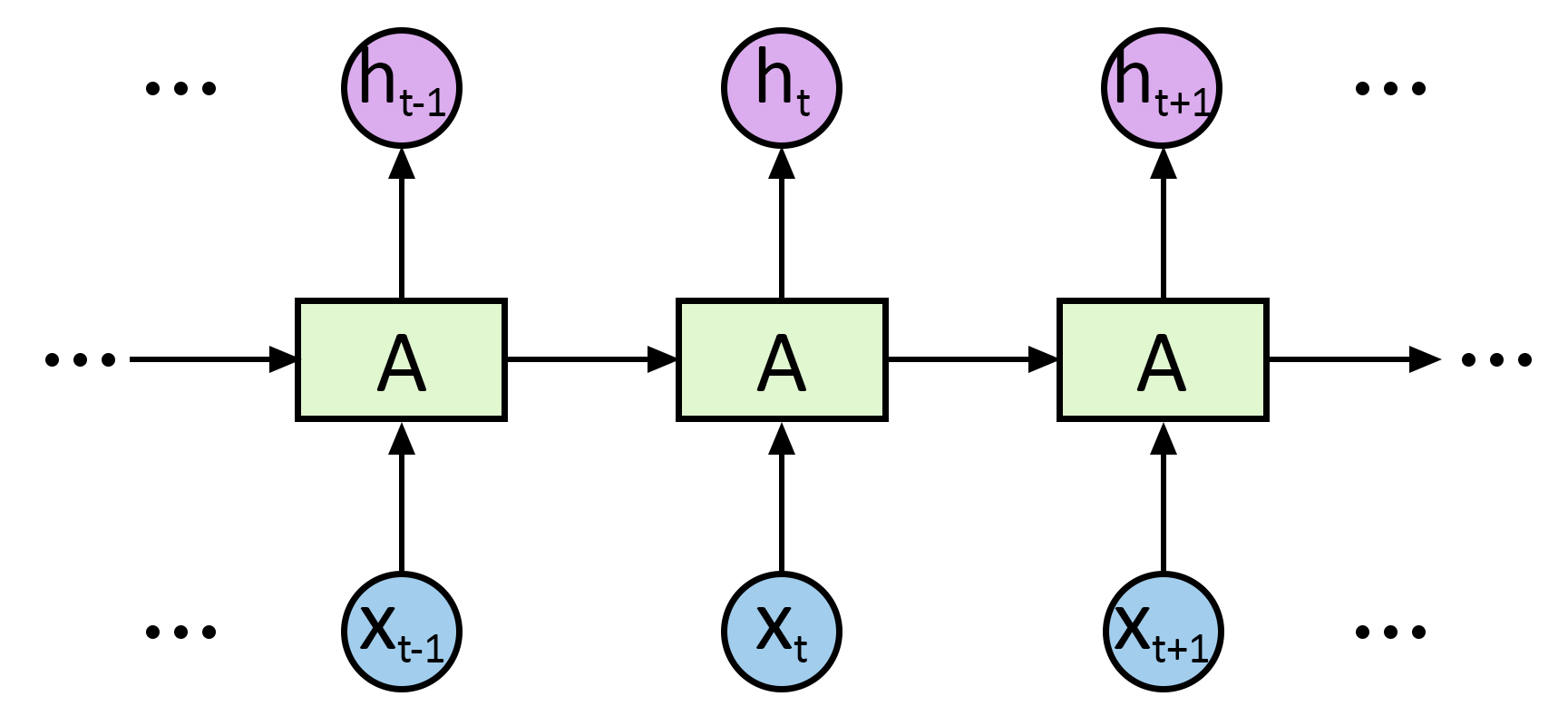}
	\caption{A simple schematic of recurrent network with the input, hidden and an output layer.}\label{fig:simpleRecurrent}
	%	\vspace{-2.0em}
\end{figure}

Recurrent Neural Networks (RNN) were proposed as a solution to overcome simple neural networks' inability to learn sequence data. In sequential data such as signals \cite{Xiong2018}, stock price \cite{Rather2015}, machine translation \cite{Liu2015}, the temporal arrangement and chain dependency of samples create meaningful patterns throughout the time. Since simple neural networks have a feed-forward structure, they cannot learn time-variant features. 
To overcome this shortcoming, different variety of feedback node connections were proposed as a variant of RNNs \cite{Chung2014, Schuster1997, Soltani2016}. These connections shape a directed graph in the temporal sequence direction that can learn and extract the sequential data's temporal instinct patterns. 
In RNNs, unlike simple neural networks, each node's output depends on the output of previous nodes. In other words, it can be said that RNNs are capable of memorizing previous computations to the current state. Fig.\ref{fig:simpleRecurrent} indicates a simple recurrent network. 

As it’s displayed in Figure \ref{fig:simpleRecurrent}, $x_t$ is the input at time step t, $y_t$ is the hidden state at step \textit{t} which is also shown as rectangle units and has the role of memory in the network, and finally $y_t$ is the hidden state at time step \textit{t}. In this manner, recurrent networks can utilize previous computations. 
Although this structure seems to be promising in terms of keep tracking of previous states and working similar to memory, simple recurrent networks are not capable of memorizing more than a few earlier time steps due to the vanishing gradient problem \cite{ShivaPrakash2019}. 
Vanishing gradient encounters when Neural Network or, in this case, RNN is being trained by gradient-based learning and backpropagation method. Backpropagation computes the gradient of the output loss with respect to the network's weights. The gradient is calculated using the chain rule and relative derivative. As a result of the consecutive multiplication of the chain rule, the gradient value usually drops to a tiny number in deeper neural networks, and as a result, the network stops learning.
This means that the network will soon be incapable of learning the complicated instinct features of the sequence data and discover the long-term dependencies. In other words, it can not remember more complicated time-dependent sequential information, which is responsible for long-term memory. 

\subsection{Long Short-term Memory }
\begin{figure}[h]
	%	\vspace{-1.0em}
	\centering
	\includegraphics[width=0.46\textwidth]{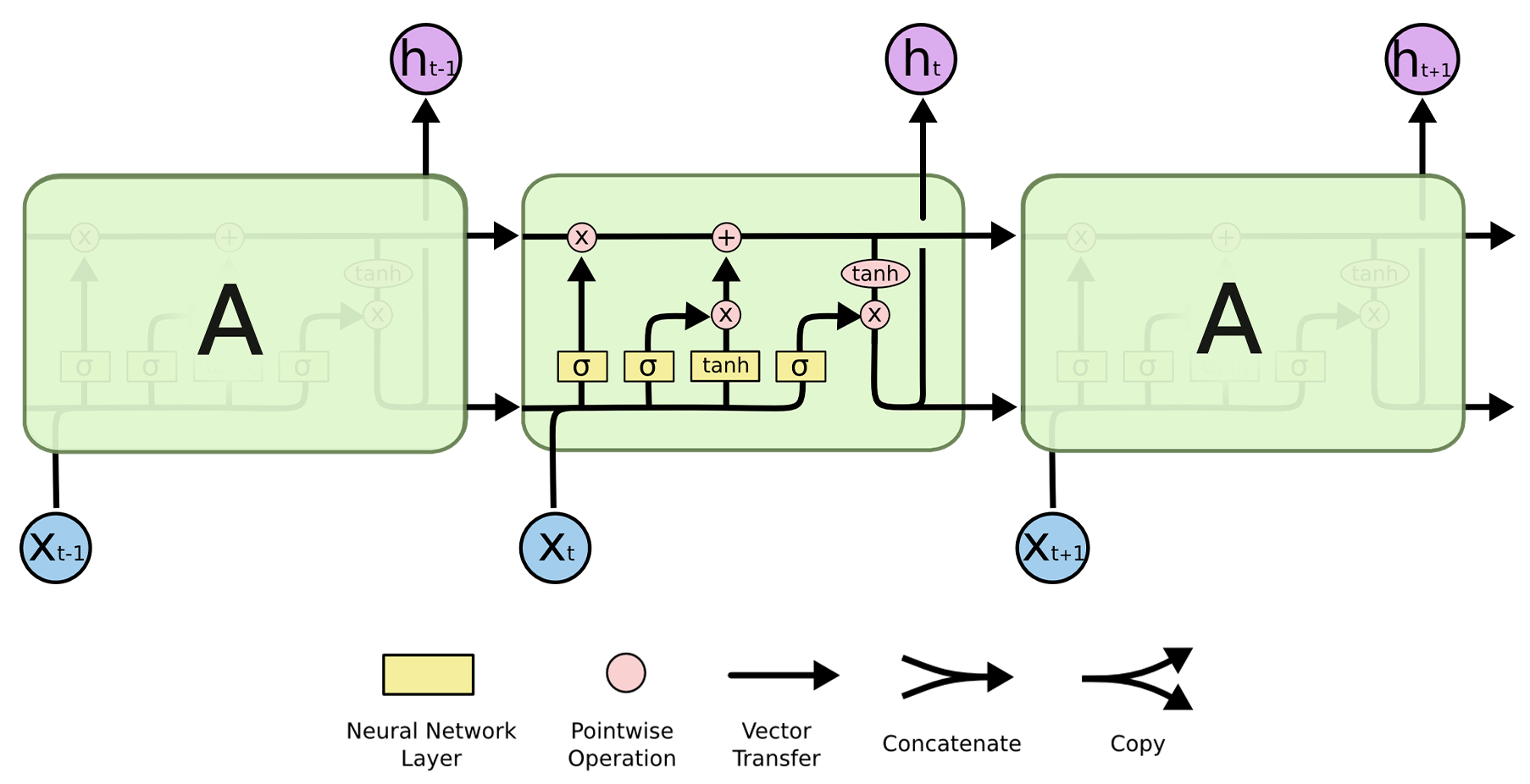}
	\caption{The overall architecture of an unrolled LSTM layer with scheme of an LSTM unit's internal structure.\\
		Image is retrieved from http://colah.github.io/posts/2015-08-Understanding-LSTMs/}\label{fig:LSTM}
	%	\vspace{-2.0em}
\end{figure}

To overcome the simple RNNs' shortcomings, Long Short-Term Memory (LSTM) \cite{Hochreiter1997} was proposed. The vanilla LSTM has the same chain-like architecture as RNN, which was introduced in the previous section. However, each memory unit of LSTM has a different structure and consists of more complicated functionalities than the vanilla RNN. 
In LSTM, each memory cell makes small modifications to the information by simple mathematical operators such as multiplication and addition on the information flowing through a mechanism called Cell states. This way, the LSTM unit can selectively keep or forget the information. 
This information generally has three main dependencies. Firstly, the previous information that is passed by the memory after the last timestep through the cell state. Secondly, the previous cell's output which is also known as the hidden state, and lastly, the input at the current timestep.
Another important term in LSTM is the analogy with conveyor belts as a mechanism to move the information flow through the LSTM block. As the information is being passed alongside the conveyor belts, the information can be added, removed, or modified by utilizing simple linear operators and Sigmoid neural net layers. This way of controlling information lets this primary component, also known as the Cell state, play a key role in keeping the main information and features for that particular time step. 
The generic architecture of LSTM is provided in Fig.\ref{fig:LSTM}.

\begin{figure*}[h]
	%	\vspace{-1.0em}
	\centering
	\includegraphics[width=\textwidth]{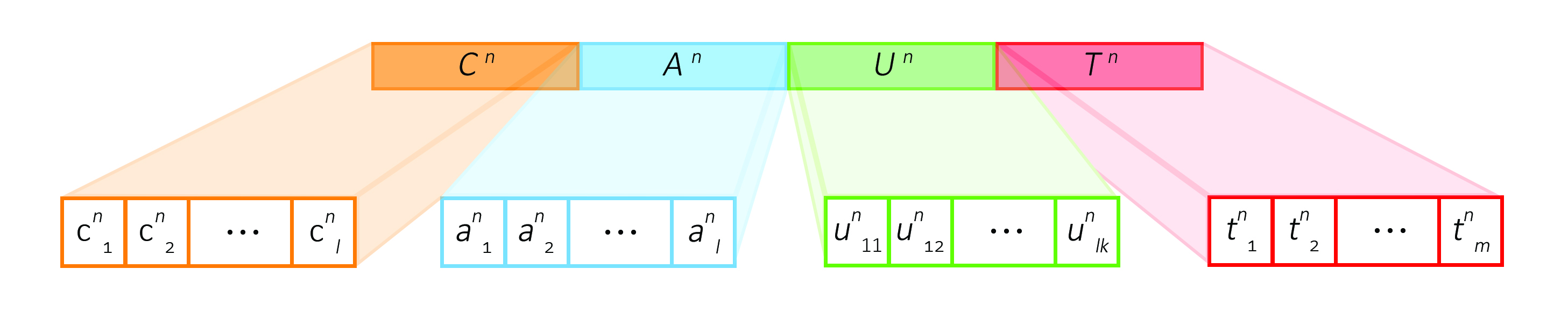}   % width=0.46\textwidth
	\caption{Each individual consists of 4 different parts with different encodings.In this image, sample individual $n$ is illustrated and splitted into the mentioned subcomponents. }\label{fig:individual}
	%	\vspace{-2.0em}
\end{figure*}

\section{Forecasting COVID-19 with NAS-BBA}
\subsection{Dataset and Challenges}
SARS-COV-2 is a newly discovered virus in late 2019. The world health organization officially announced the pandemic caused by this virus on December 31st, 2019. Therefore, to study this virus's epidemiological behavior, especially in the first states of its outbreak, there was not much data available to analyze. It's also worth mentioning that, to have a fair epidemiological evaluation and train an accurate model of the pandemic, we must only study the regions with the same culture and social behavior since an epidemic is highly dependent on those factors. These reasons lead us to have very limited data.
In this paper, we use the open geographic distribution data of COVID-19 cases worldwide retrieved from the European Centre for Disease Prevention and Control\footnote[1]{\url{https://www.ecdc.europa.eu/en/geographical-distribution-2019-ncov-cases}} to build a model for forecasting Iran's daily cases on non-lockdown days. The raw version of this dataset consists of 12 features. An overview of 5 samples of the data with some of their main features is provided in Table .\ref{tab:rawdata}. We utilize two features from this dataset to combine with two new features that we will introduce shortly. 

\begin{table*}[!htb]
	\captionsetup{size=footnotesize}
	\caption{An overview of several features from the raw data.} \label{tab:rawdata}
	\setlength\tabcolsep{0pt} % let LaTeX compute intercolumn whitespace
	\footnotesize\centering
	European Centre for Disease Prevention and Control’s Dataset
	
	\smallskip 
	\begin{tabular*}{\textwidth}{@{\extracolsep{\fill}}cccccccc}
		\toprule
		dateRep  & day & month  & year & cases & deaths & countriesAndTerritories & Cumulative\_number  \\ 
		\midrule  \vspace{2pt} 
		20/03/2020 & 20 & 3 & 2020 & 1046 & 149 & Iran & 17.963 \\ \vspace{2pt} 
		21/03/2020 & 21 & 3 & 2020 & 1237 & 146 & Iran & 17.968 \\ \vspace{2pt} 
		22/03/2020 & 22 & 3 & 2020 & 966 & 123 & Iran & 17.834 \\ \vspace{2pt} 
		23/03/2020 & 23 & 3 & 2020 & 1028 & 129 & Iran & 18.177 \\ \vspace{2pt} 
		24/03/2020 & 24 & 3 & 2020 & 1411 & 127 & Iran & 19.162 \\

		\bottomrule
	\end{tabular*}
\end{table*}
Since one of the main causes of spreading COVID-19 is human interactions and in-person communications, controlling this matter was one of the macro decision-makers first concerns. As a result, by early April 2020, over one-third of the global population was under some form of movement restriction, quarantine, or COVID-19 lockdown. Although to prevent further economic damage, most countries' health organizations started considering new protocols for routine economic activities and workplaces such as decreasing the number of employees and monitoring their health condition \cite{Cirrincione2020}. Meanwhile, research works \cite{Tay2020} showed that there is another important factor that can have a serious impact on the outbreak despite all the protocols, and that is the tendency of people to break the quarantine and have in-person social communications \cite{Koh2020}. Therefore one of the main factors we have taken into account in this paper is the impact of non-workdays or holidays on the pandemic case numbers.

To do so, we introduce the first augmented feature by determining the type of days based on whether it's a holiday or it's a regular workday and is called "d\_type". We extracted the holidays' status of Iran from Google Calendar API and gave them the value of 1 if the corresponding day was a holiday and 0 if it's a workday. The second augmented feature is extracted from the holidays feature because each holiday increases people's tendency for unnecessary gatherings in quarantine. Therefore we introduced the "gathering" feature, and each sample gets a value of 1 if it's a holiday or it's a non-holiday, and between two holidays; otherwise, it gets 0. Lastly, we use an index to keep track of the sequence. A part of the new data is shown in Table .\ref{tab:AugmentedData}. 

\begin{table}[h]
	%	\vspace{-1.5em}
	\renewcommand{\arraystretch}{1.1}
	%\small
	\tabcolsep=0.08cm
	\begin{center}
		\caption{An overview of the indexed new data with feature augmentation.}\label{tab:AugmentedData}
		\vspace{0em}
		\scalebox{1} {
			\begin{tabular}{ccccc} % m{1.64cm}m{1.64cm}m{1.64cm}m{1.64cm}m{1.64cm}
				\hline
				index & \hspace{1.1em}cases & \hspace{1.1em}c\_num & \hspace{1.1em}d\_type & \hspace{1.1em}gathering \\ 
				
				%	\multicolumn{1}{c}{index} & \multicolumn{1}{c}{cases} & \multicolumn{1}{c}{Cumul\_num} & \multicolumn{1}{c}{d\_Status} & \multicolumn{1}{c}{g\_possibility}\\
				\hline
				128 &\hspace{1.1em} 2472 &\hspace{1.1em} 43.371 &\hspace{1.1em} 0 &\hspace{1.1em} 0  \\
				129 &\hspace{1.1em} 2449 &\hspace{1.1em} 42.732 &\hspace{1.1em} 0 &\hspace{1.1em} 0 \\
				130 &\hspace{1.1em} 2563 &\hspace{1.1em} 42.064 &\hspace{1.1em} 1 &\hspace{1.1em} 1  \\
				131 &\hspace{1.1em} 2612 &\hspace{1.1em} 41.434 &\hspace{1.1em} 0 &\hspace{1.1em} 1 \\
				132 &\hspace{1.1em} 2596 &\hspace{1.1em} 40.255 &\hspace{1.1em} 1 &\hspace{1.1em} 1 \\
				\hline
			\end{tabular}
		}
	\end{center}
	\vspace{-0.0em}
\end{table}

\subsection{Neural Architecture Search with BBA}
To deal with the mentioned challenges, we have taken an Evolutionary Neural Architecture Search (NAS) approach to optimize the deep model hyperparameters. The optimization of hyperparameters is an NP-hard problem, which means finding the optimal solution we require to perform an exhaustive search in the solution space using metaheuristic techniques. Therefore we have chosen the BBA as it is a widely used metaheuristic algorithm \cite{Gupta2019,Nakamura2012} and as the main paper claims, it is superior to its other competitive binary algorithms. From now, we refer to this proposed framework as NAS-BBA.

Previous research \cite{Zoph2016,Stanley2002} suggest that neural architecture search techniques are able to design the simplest topology for the network as well as increase the performance of the final output. The NAS approach also develops a deep architecture with a sufficient number of parameters and not too many. This helps the model deal with limited data and learn abstract information from layers without getting overfitted. In the meanwhile, training time is longer for RNNs architecture compared to the architectures that can process the data in parallel. Moreover, adjusting the numbers of LSTM layers and the number of units in each layer will result in a large number of architectures. This is a time-consuming process and requires so much trial and error. Therefore NAS is a handy and reasonable approach to design an efficient deep model.

Before using BBA to optimize the model, there are two main factors that we have to consider. We first have to define an encoding for the population's individuals so it can represent the problem clearly. The second thing that we have to focus on is utilizing a convenient fitness function for the problem. We will discuss these two factors and how we customized them for forecasting COVID-19 cases.
\subsubsection{Defining Individuals}

The individuals are defined using a hybrid encoding structure as the population of BBA. Each individual consists of 4 parts. The first two $C_l^n$  and $A_l^n$ encoded in the Binary scheme. Vector $C_l^n$ is responsible for determining the existence of a layer. In other words, if element $C_3^n$ has the value of 1, it means the layer is activated, and if it has the value of 0, it shows the absence of the corresponding layer in the model. The second binary vector $A_l^n$ represents the activation function used in each layer. In this study, we encoded ReLU with 1 and Sigmoid function by 0.  
The last two vectors are encoded in gray-code. The third vector is $U_k^n$ that can be split into \textit{k} subvectors. Each of the \textit{k}  new vectors determine the number of units in the corresponding layer, and lastly, the fourth vector $T^n$ represents the number of timesteps in which we use to frame the data for the sequential model. A simple representation of this encoding scheme is provided in Figure.\ref{fig:individual}.
The overall number of elements in each individual is fixed and can be calculated as Eq.\eqref{eq:Eq.9}. $l$ and $a$ are the maximum numbers of layers and activations, respectively. For instance, $l=3$ means that there are three layers that BBA can determine their existence. The first logarithm term in the equation is responsible for converting the maximum number of units in each layer $k$ to the suitable number of binary units capable of representing it. Likewise, the second logarithm term converts the maximum timesteps that we defined to the number of gray-code encoding units. It is also worth mentioning that $\left\lceil x\right\rceil$ maps $x$ to the least integer, greater than or equal to $x$.

\begin{equation} \label{eq:Eq.9}
	L=l+a+\sum_{k=1}^{l}\left\lceil\log_2{u_k}\right\rceil+\ \left\lceil\log_2{t}\right\rceil
\end{equation}

\subsubsection{Selecting Fitness Function}
For evaluating each Deep Model corresponding to each individual in the population, we need to select a convenient fitness function. Since the final goal is forecasting COVID-19 daily cases, we can conclude that the problem is regression. As the literature of artificial neural networks and deep learning models in regression problems suggests, we select Mean Squared Error (MSE) as BBA's fitness function. Eq.\eqref{eq:Eq.10}

\begin{equation} \label{eq:Eq.10}
	MSE=\ \frac{1}{n}\sum_{i=1}^{n}{(y_i-{\hat{y}}_i)^{2}}
\end{equation}

\subsubsection{Training }
In the training phase, for each generation, the current population will be altered as in Eq.\eqref{eq:Eq.8}. Each individual will then be split into parts explained previously and mapped to the corresponding component of the deep model as a candidate solution. Then the deep model will be trained with training as long as the specified criteria are met. Finally, the trained model will be evaluated with unseen data, and the MSE value will be returned to the BBA as the individual's fitness value. In this way, we can evaluate every generated model. This process will end when BBA's termination condition is reached, and the $gbest$ will be returned as the best solution. 
\section{Experiments}
In this section, we first introduce the deep structure used for forecasting COVID-19 cases, then we address the experimental setting and specified parameters, and finally, we discuss the experimental results.
\subsection{Deep Model Structure}
\begin{figure}[h]
	%	\vspace{-1.0em}
	\centering
	\includegraphics[width=0.46\textwidth]{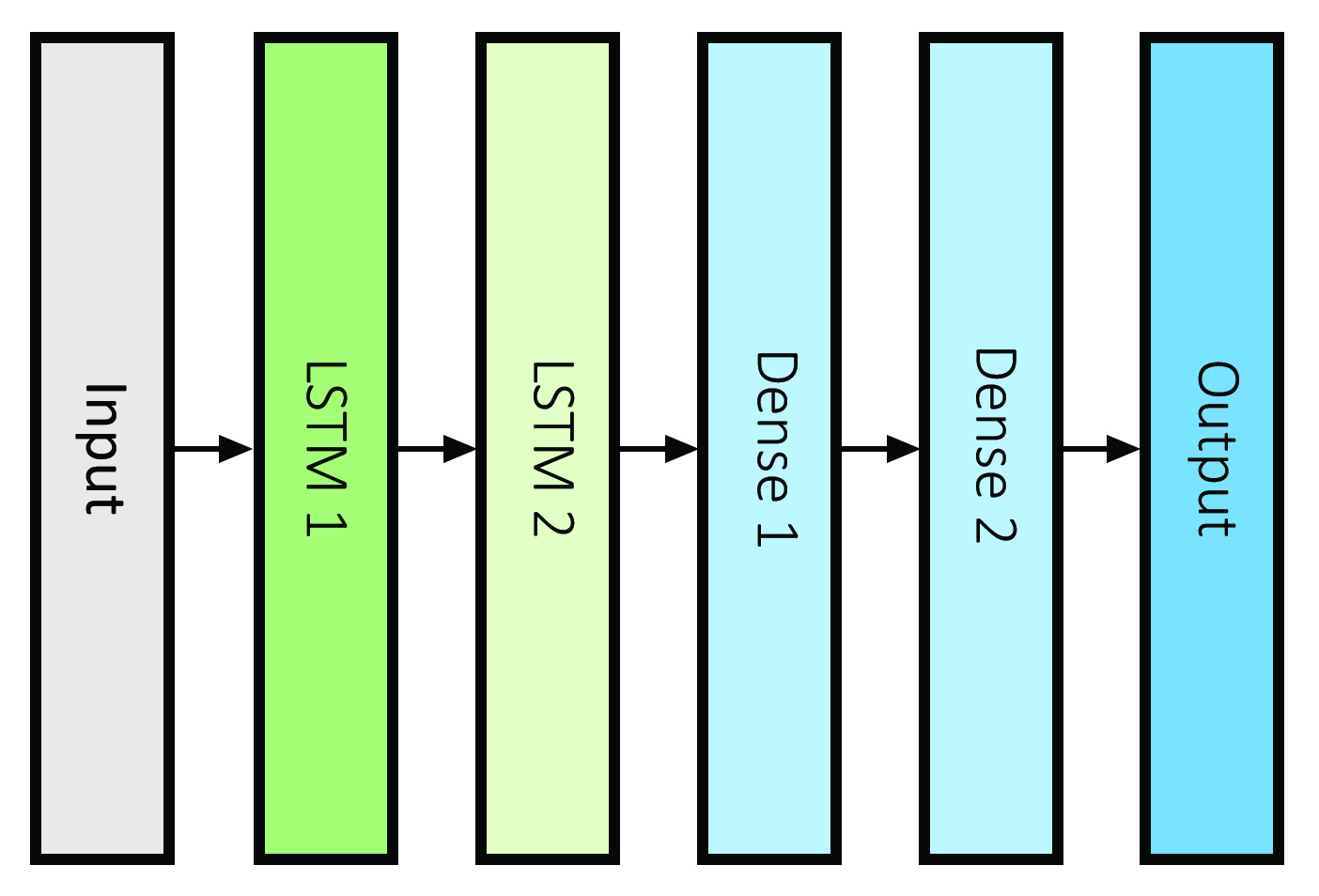}
	\caption{An overview of the whole deep model structure. LSTM 1 shown in dark green and the Output layer in dark blue show the fixed layers in the model.}\label{fig:DeepStructure}
	%	\vspace{-2.0em}
\end{figure}

In this paper, we utilized a 5-layer deep recurrent network using vanilla LSTM units to forecast COVID-19 cases. We used an architecture consisting of two LSTM layers and two dense layers, and the output layer. Since we need at least one LSTM layer to extract the time dependency information and one output layer to output the predicted result, we consider the first and last layers as fixed layers and do not define a $C_l^n$ element in the individual corresponding to these layers. A simple scheme of this architecture is illustrated in Fig.\ref{fig:DeepStructure}.

\subsection{Experimental Setting}
\begin{itemize}
	
	\item Dataset: In the experiments, we determined the maximum number of timesteps for framing the sequence data to 31. This number requires 5 elements of our individuals to be encoded in gray code. Before the evaluation of each individual, data is first framed into $n$ samples of $t$ timesteps and $f$ features as shown in Fig.\ref{fig:SequenceTensor}. We split it into two train and test data with the ratio of  80:20, respectively. Then we normalize the data, so it gets rescaled to the range of $\left[0,1\right]$ and use the train data for the training phase and test data to evaluate the model.\vspace{5pt}
	
	\begin{figure}[h]
		%	\vspace{-1.0em}
		\centering
		\includegraphics[width=0.46\textwidth]{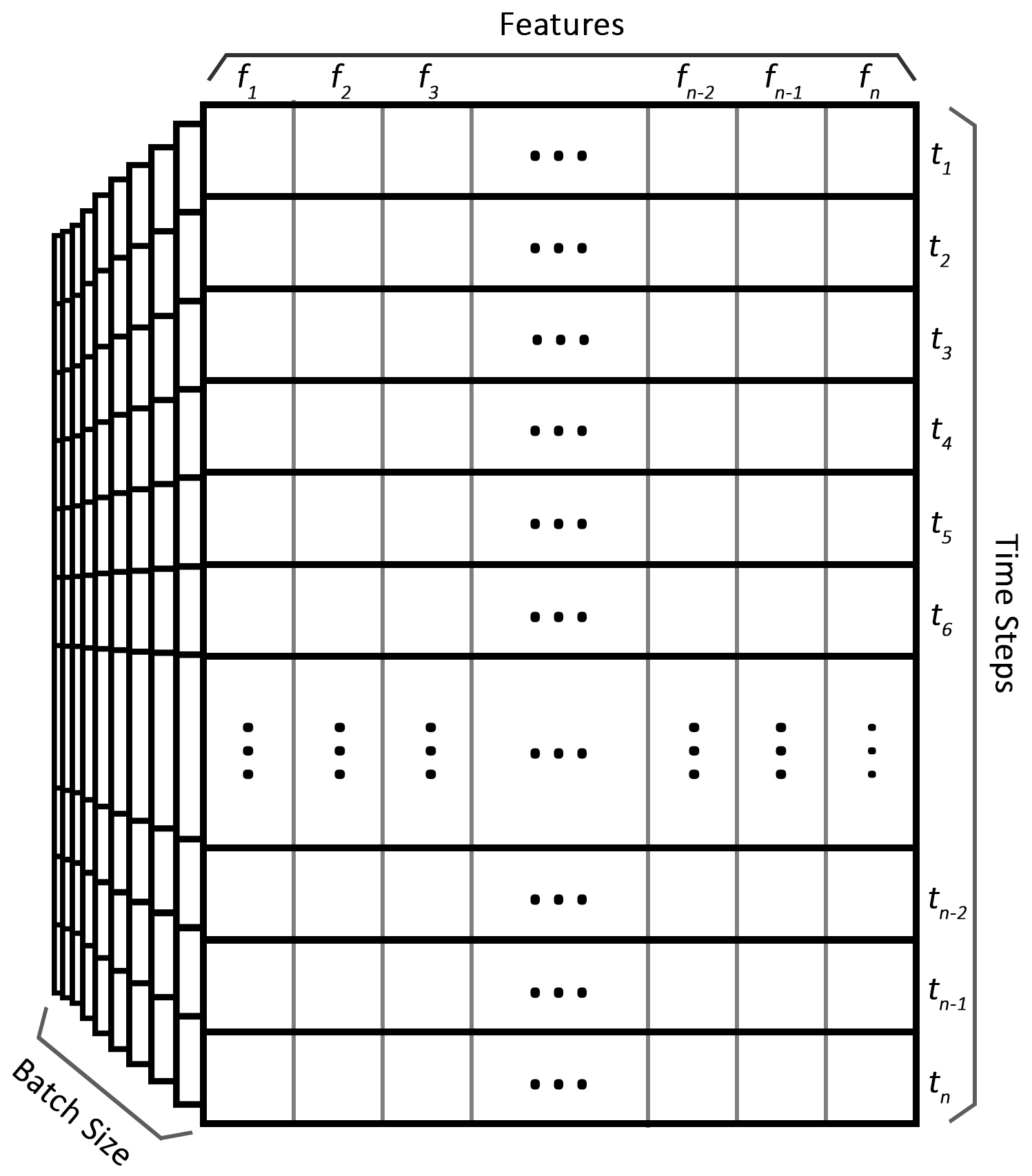}
		\caption{A schematic overview of $n$ samples with $t$ timesteps and $f$ features.}\label{fig:SequenceTensor}
		%	\vspace{-2.0em}
	\end{figure}
	
	\item Deep Model: For the deep model, the maximum number of units in each layer is 31 for each normal LSTM layer and 63 for the other two dense layers. For the last layer, we used a single neuron to predict the output. As the literature suggests, we added a dropout rate of 0.8 to each LSTM layer and l2 regularization with a lambda rate of 0.01. We encoded the ReLU function by 1, and the Sigmoid function by 0 for each individual's activation element corresponding to dense and outputs layers' activations. The final structure of each individual consists of 32 elements. In the experiments, we trained each model by 200, 500, and 1000 epochs to evaluate the BBA's individuals' fitness, and after obtaining the best architecture, we trained the model by 2,000 epochs. Throughout this study, we run every experiment three times and report the mean RMSE loss as the final score. 
	\vspace{5pt}
	\item BBA: The number of population, iterations and input parameters of the BBA is set as the base research paper \cite{Mirjalili2014} determined and is provided in Table \ref{tab:BBASetting}. It should be mentioned that, due to the high computational time of fitness evaluation, the number of BBA's iteration, population number, and Deep Models' epochs are kept limited for the experiments.
\end{itemize}

\begin{table}[h]
	%	\vspace{-1.5em}
	\renewcommand{\arraystretch}{1.1}
	%\small
	\tabcolsep=0.08cm
	\begin{center}
		\caption{BBA Experimental settings.}\label{tab:BBASetting}
		\vspace{0em}
		\scalebox{1} {
			\begin{tabular}{lc} % m{1.64cm}m{1.64cm}m{1.64cm}m{1.64cm}m{1.64cm}
				\hline
				Parameters \hspace{1.3em} & \hspace{1.1em}Values \\ 
				
				%	\multicolumn{1}{c}{index} & \multicolumn{1}{c}{cases} & \multicolumn{1}{c}{Cumul\_num} & \multicolumn{1}{c}{d\_Status} & \multicolumn{1}{c}{g\_possibility}\\
				\hline
				
				Population\hspace{1.3em} &\hspace{1.1em} 10,30  \\
				$F_min$ \hspace{1.3em} &\hspace{1.1em} 0  \\
				$F_max$ \hspace{1.3em} &\hspace{1.1em} 1  \\
				A  \hspace{1.3em} &\hspace{1.1em} 0.25 \\
				r \hspace{1.3em} &\hspace{1.1em} 0.5 \\ 
				$\epsilon$ \hspace{1.3em} &\hspace{1.1em} [-1,1]\\
				$\alpha$ \hspace{1.3em} &\hspace{1.1em} 0.9 \\
				$\gamma$ \hspace{1.3em}  &\hspace{1.1em} 0.9 \\
				BBA iterations \hspace{1.3em}  &\hspace{1.1em} 100\\
				\hline
				Model iterations \hspace{1.3em}  &\hspace{1.1em} 200, 500, 1000\\
				
				\hline
			\end{tabular}
		}
	\end{center}
	\vspace{-0.0em}
\end{table}

The model is implemented with Tensorflow 2.2.0 in Python bound with BBA's MATLAB code retrieved from https://www.mathworks.com/matlabcentral/fileexchange/44707-binary-bat-algorithm. To obtain unbiased results, all experiments are conducted using the same PC with the detailed configuration settings, as shown in Table \ref{tab:PCSetting}.

\begin{table}[h]
	%	\vspace{-1.5em}
	\renewcommand{\arraystretch}{1.1}
	%\small
	\tabcolsep=0.08cm
	\begin{center}
		\caption{Experimental environment's configuration settings.}\label{tab:PCSetting}
		\vspace{0em}
		\scalebox{1} {
			\begin{tabular}{lc} % m{1.64cm}m{1.64cm}m{1.64cm}m{1.64cm}m{1.64cm}
				\hline
				Name & \hspace{1.05em}Configuration Settings  \\ 
				
				%	\multicolumn{1}{c}{index} & \multicolumn{1}{c}{cases} & \multicolumn{1}{c}{Cumul\_num} & \multicolumn{1}{c}{d\_Status} & \multicolumn{1}{c}{g\_possibility}\\
				\hline
				\small \vspace{1.05em}\textit{\footnotesize\textbf{Hardware}} &    \\ 
				\footnotesize CPU &\footnotesize\hspace{1.05em} Intel Core i7-6700HQ  \\
				\footnotesize CPU Frequency &\footnotesize\hspace{1.1em} 2.60GHz  \\
				\footnotesize RAM &\footnotesize \hspace{1.05em} 32GB  \\
				\footnotesize GPU &\footnotesize\hspace{1.05em} NVIDIA GeForce GTX 980\\ 
				\hline 
				\small \vspace{1.05em}\footnotesize\textit{\textbf{Software}} &     \\
				\footnotesize Operating System &\footnotesize\hspace{1.05em} Windows 10 Pro 64-bit  \\
				\footnotesize Python &\footnotesize\hspace{1.05em} 2.7.6  \\
				\footnotesize Implementation Environment &\footnotesize\hspace{1.05em} MATLAB R2018b  \\
				\footnotesize Tensorflow &\footnotesize\hspace{1.05em} 2.2.0  \\
				\hline
			\end{tabular}
		}
	\end{center}
	\vspace{-0.0em}
\end{table}

\subsection{Experimental Results}
To evaluate the effectiveness of the proposed approach, we conducted several different experiments on the COVID-19 dataset. We first run the framework on a population of 10 and 20 with 200 epochs and compare the two output models. Then we studied the influence of epoch number on the improvement of the final architecture by setting it to 200, 500, and 1000 (M1-M3) and compared them with five customized models (Network1-Network5). 
To study the introduced data features' effectiveness, we train and test the best model on the initial data and the new data with augmented features(M1 vs. M4). The obtained architectures from the NAS-BBA framework and the self-defined architectures with their corresponding detailed information are provided in Table .\ref{tab:ModelComparisons}. 
\subsubsection{Results}
\begin{figure}[h]
	%	\vspace{-1.0em}
	\centering
	\includegraphics[width=0.46\textwidth]{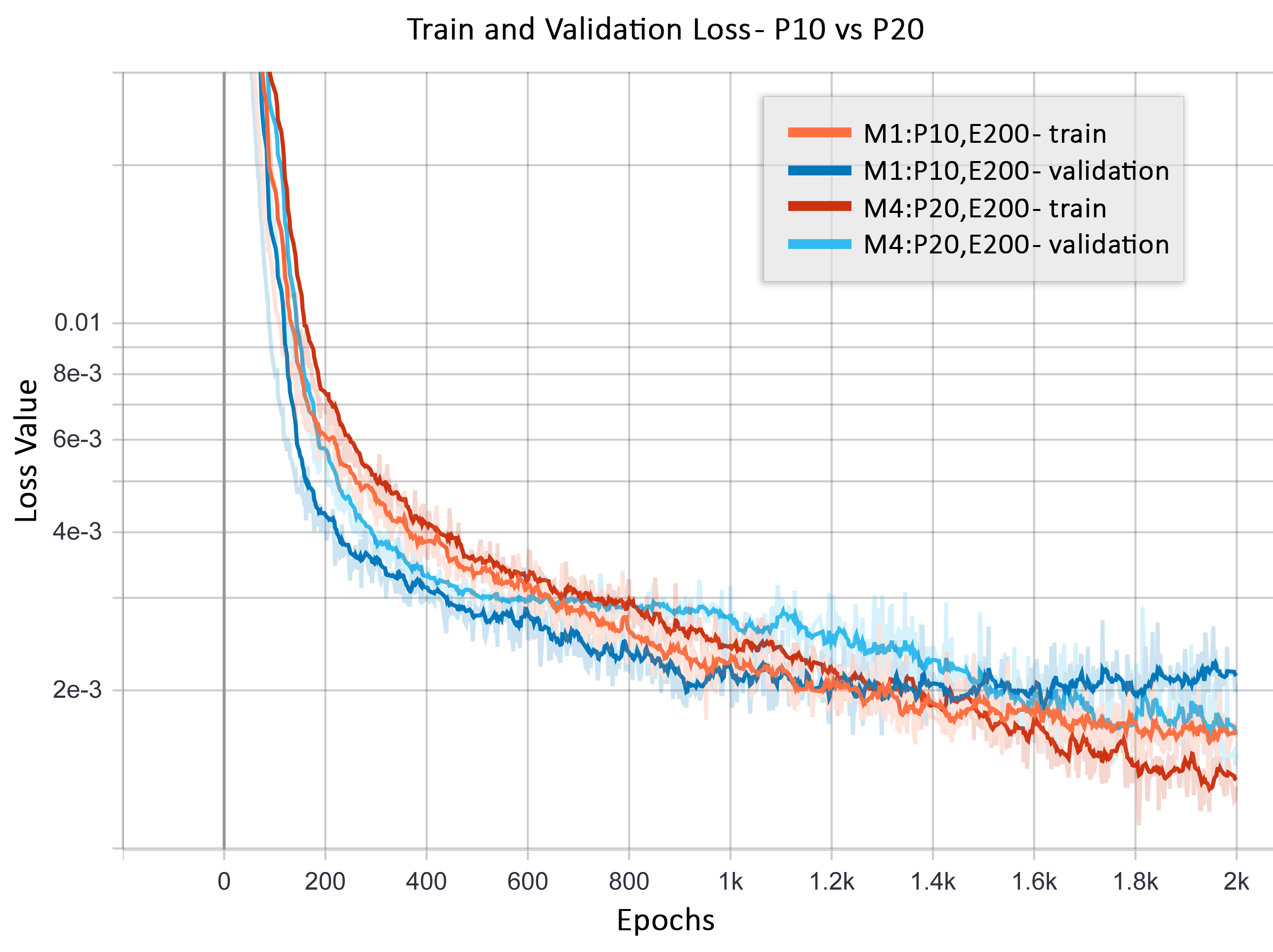}
	\caption{Train and Validation loss value of the model obtained by NAS-BBA with 10 and 20 individuals with 100 iteration and 200 epochs.}\label{fig:p10vsp20}
	%	\vspace{-2.0em}
\end{figure}

As it can be observed in Table.\ref{tab:ValidationLoss}, there was a meaningful improvement in M1 to M3 networks when the epoch numbers increased from 200 to 1000. This also proves that the higher numbers of epochs give a sufficient amount of time to the NAS-BBA framework for a better evaluation of fitness corresponding to each individual. In other words, suppose we set the number of epochs to 1000 for the framework. This helps the deep architecture corresponding to each individual to be trained for a longer time and, as a result, provide a more accurate RMSE as fitness value, and therefore the best individual will be chosen with less error. 
Also, to study the effect of the population number on the framework's accuracy, we conducted experiments on NAS-BBAS with 10 and 20 individuals (M1 and M4). Due to the high computational time of Deep Models evaluations for each individual, we kept the number of epochs to 200 for the BBA fitness evaluation. As shown in Fig.\ref{tab:ModelComparisons}, the mean loss value obtained by the NAS-BBA with 20 individuals had significant improvement compared to the 10-individual version. It is also evident in Fig.\ref{fig:p10vsp20} that both validation and train loss of NAS-BBA with a population of 20 (P20) has a decreasing trend to the last epoch. On the contrary, the P10 version almost started getting overfitted from the 1700th epoch, and the validation loss started increasing from then. 

\begin{table}[h]
	%	\vspace{-1.5em}
	\renewcommand{\arraystretch}{1.1}
	%\small
	\tabcolsep=0.08cm
	\begin{center}
		\caption{Mean RMSE loss obtained by 3 individual runs over COVID-19 dataset with augmented features.}\label{tab:ValidationLoss}
		\vspace{0em}
		\scalebox{1} {
			\begin{tabular}{lc} % m{1.64cm}m{1.64cm}m{1.64cm}m{1.64cm}m{1.64cm}
				\hline
				Model & \hspace{3.1em}Mean RMSE Loss  \\ 
				
				%	\multicolumn{1}{c}{index} & \multicolumn{1}{c}{cases} & \multicolumn{1}{c}{Cumul\_num} & \multicolumn{1}{c}{d\_Status} & \multicolumn{1}{c}{g\_possibility}\\
				\hline
				M1:P10,E200 &\hspace{3.1em} 1.85e-3    \\
				M2:P10,E500 &\hspace{3.1em} 1.61e-3   \\
				M3:P10,E1000 &\hspace{3.1em} 1.35e-3    \\
				M4:P20,E200 &\hspace{3.1em} \textbf{1.23e-3}    \\
				Network1 &\hspace{3.1em} 3.39e-3 \\
				Network2 &\hspace{3.1em} 4.50e-3 \\
				Network3 &\hspace{3.1em} 6.69e-3 \\
				Network4 &\hspace{3.1em} 2.74e-3 \\
				Network5 &\hspace{3.1em} 3.15e-3 \\
				\hline
			\end{tabular}
		}
	\end{center}
	\vspace{-0.0em}
\end{table}

\begin{table*}[!htb]  %!htb
	\captionsetup{size=footnotesize}
	\caption{This table provides the main hyperparameteres that represent each architecture used in the experiments. The M1 to M4 models are obtained from the NAS-BBA framework and Network1 to Network 5 are self-defined architectures  for better evaluation of the models.} \label{tab:ModelComparisons}
	\setlength\tabcolsep{0pt} % let LaTeX compute intercolumn whitespace
	\footnotesize\centering
	Deep Models' Architecture Used for Forecasting COVID-19 Cases
	
	\smallskip 
	\begin{tabular*}{\textwidth}{@{\extracolsep{\fill}}lccccccc}
		\toprule
		Network Name  & Time steps & Existence & Activation Functions & LSTM1 & LSTM2  & Dense1 & Dense2   \\ 
		\midrule  \vspace{2pt} 
		M1:P10,E200 &  21 & EEE & RR & 18 & 26 & 9 & 63 \\ \vspace{2pt} 
		M2:P10,E500 &  16 & EEE & RR & 24 & 27 & 16 & 3 \\ \vspace{2pt} 
		M3:P10,E1000 & 23 & EEE & RR & 12 & 29 & 16 & 2 \\ \vspace{2pt} 
		M4:P20,E200 & 24 & EEE & RR & 25 & 20 & 9 & 33  \\ \vspace{2pt} 
		Network1 &  32 & EEE & RR & 32 & 32 & 64 & 64 \\ \vspace{2pt} 
		Network2 &  28 & EEE & RR & 20 & 20 & 32 & 32 \\ \vspace{2pt} 
		Network3 &  20 & EEE & RR & 20 & 20 & 32 & 32 \\ \vspace{2pt} 
		Network4 &  16 & EEE & RR & 24 & 24 & 16 & 32 \\ \vspace{2pt} 
		Network5 &  10 & EEE & RR & 16 & 16 & 16 & 32 \\ %\vspace{2pt} 
		\hline
		%	\cline{1-9}
		\multicolumn{2}{l}{\scriptsize E: Existent } & \multicolumn{2}{l|}{\scriptsize N: Non-Existent:} & \multicolumn{2}{l}{\scriptsize R: ReLU  } & \multicolumn{2}{l}{\scriptsize S: Sigmoid}\\		
		\bottomrule
	\end{tabular*}
\end{table*}

\begin{figure}[h]
	%	\vspace{-1.0em}
	\centering
	\includegraphics[width=0.46\textwidth]{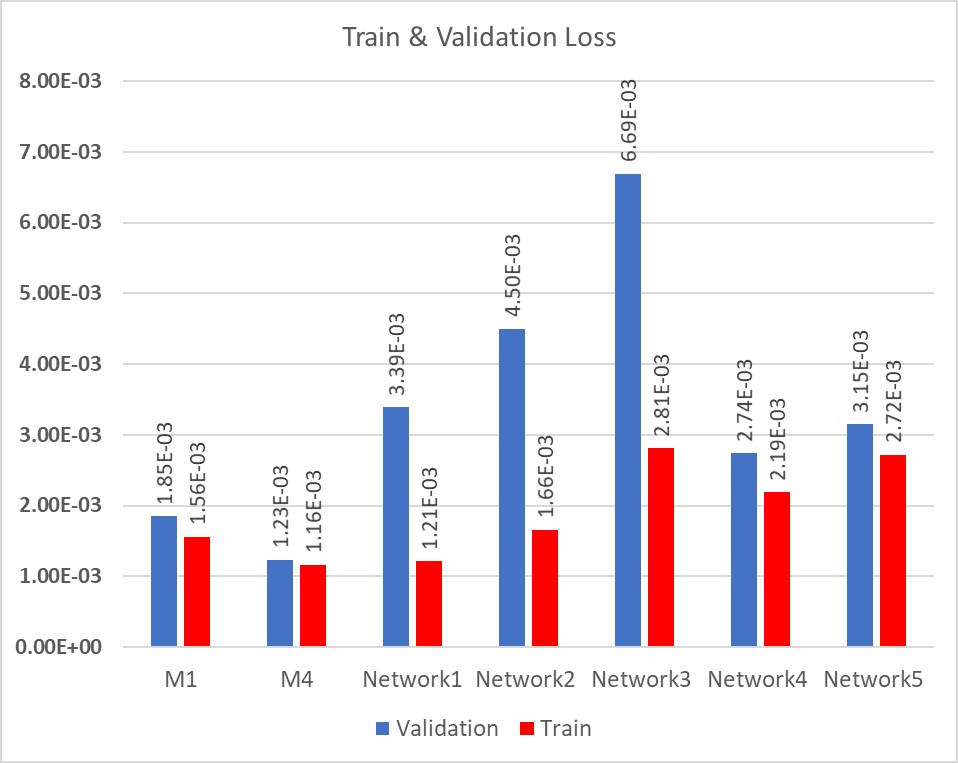}   % width=0.46\textwidth
	\caption{Train and validation loss of the main architectures.}\label{fig:LossChart}
	%	\vspace{-2.0em}
\end{figure}

\begin{figure*}[h]
	%	\vspace{-1.0em}
	\centering
	\includegraphics[width=\textwidth]{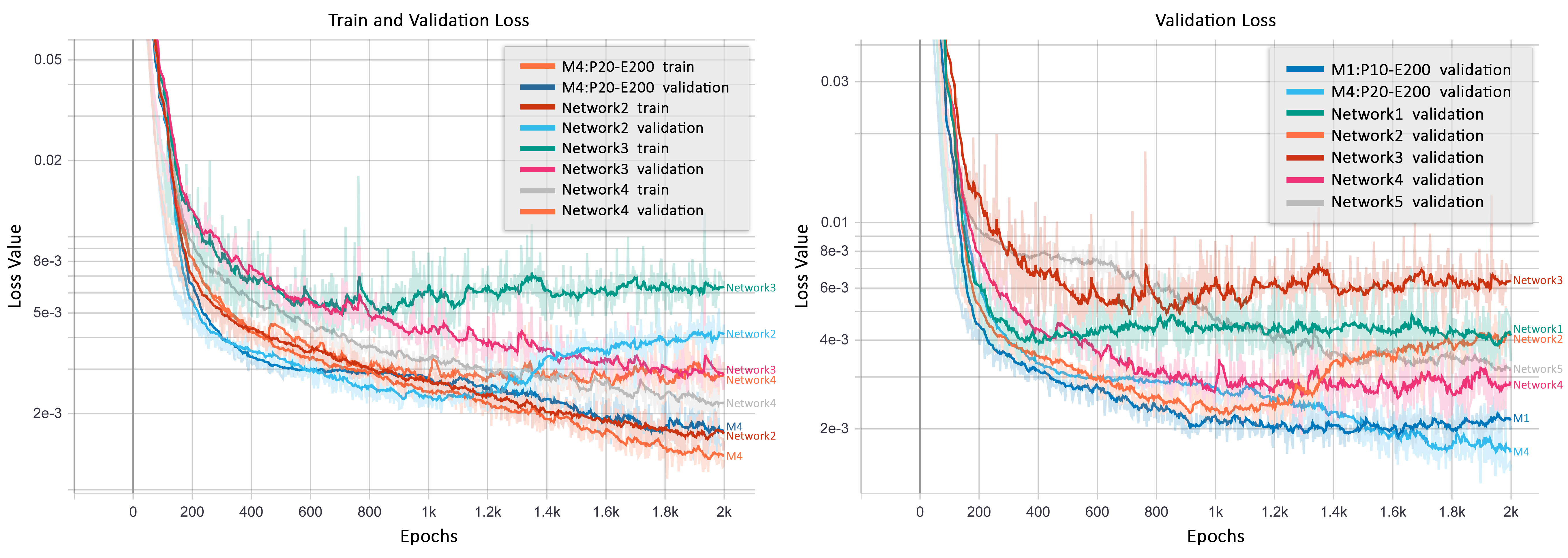}   % width=0.46\textwidth
	\caption{The left graph shows the train and validation Loss graphs of the most important models and the right graphs displays the Loss value on only the validation data.}\label{fig:BigAll}
	%	\vspace{-2.0em}
\end{figure*}

To further show the importance of having an optimized architecture to forecast COVID-19 cases and also showing the effectiveness of NAS-BBA, the performance of 5 more networks (Network1-Network5) was evaluated. We introduce these networks by setting their hyperparameters in the initially defined range. The important train and validation loss graphs of models are provided in Fig.\ref{fig:BigAll}, and for better observing differences, train and validation's loss values are plotted in Fig.\ref{fig:LossChart}.

From the left learning curve graph of the Network2 model in Fig.\ref{fig:BigAll}, it is evident that although the training learning curve of this model keeps a decreasing trend till the last epoch, the validation curve starts getting a sharply increasing curve after around the 1000th epoch. On the other side, it can also be seen that Network4's train and validation loss graph both keep the decreasing trend almost through the whole training phase, but the network cannot decrease the loss value from the 1600th epoch. This happens due to the insufficient number of time steps or hidden units. Lastly, we can see that Network3 also gets overfitted shortly after around epoch 700 and doesn't have any further improvements despite the fact that its hyperparameters such as hidden units and timesteps are closer to the one selected by NAS-BBA.

\begin{figure}[h]
	%	\vspace{-1.0em}
	\centering
	\includegraphics[width=0.46\textwidth]{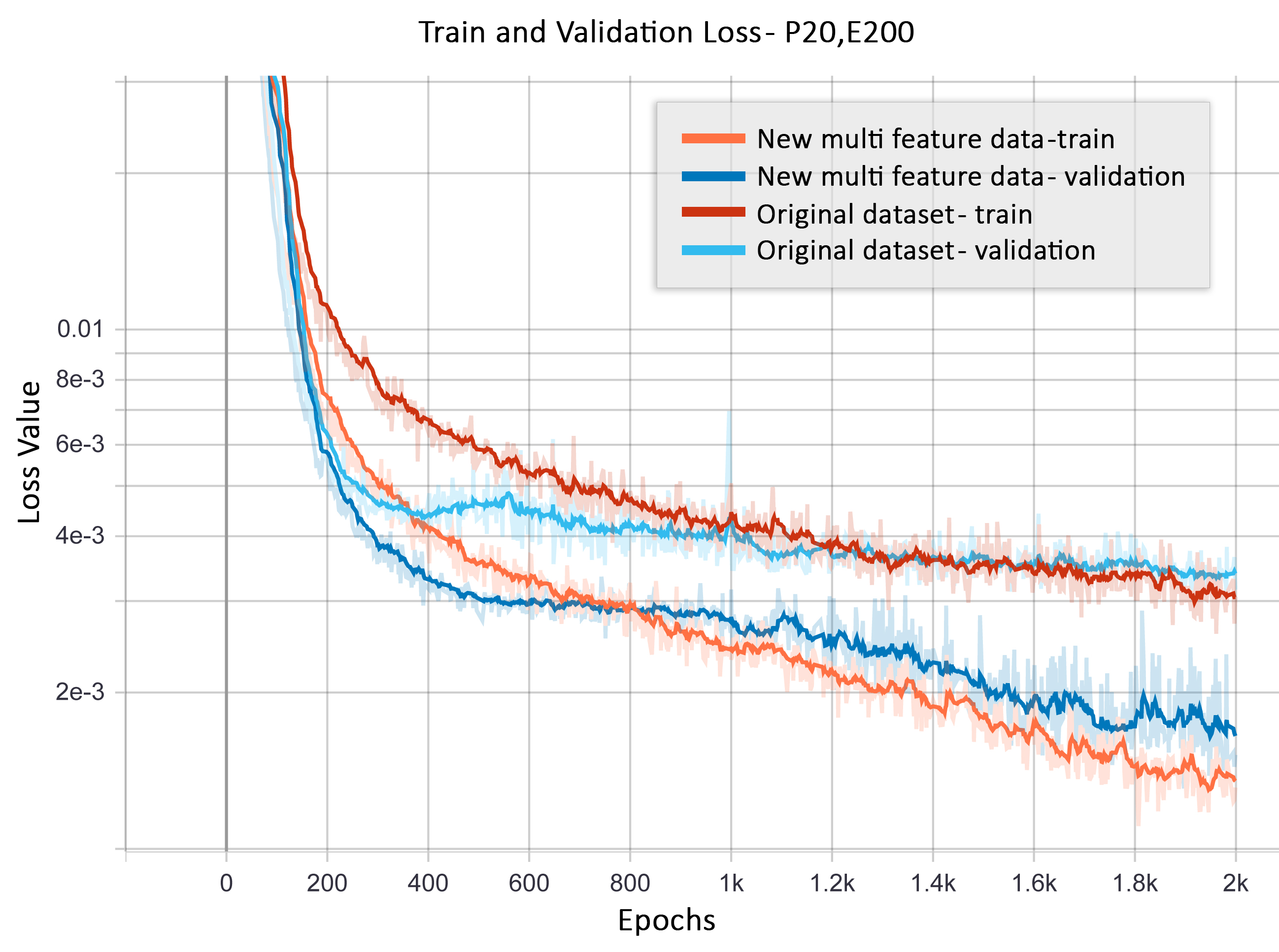}   % width=0.46\textwidth
	\caption{Learning curve of M4 architecture, showing the loss of training and validation on the new COVID-19 dataset with augmented features and the original data.}\label{fig:Epoch_Loss_Two_Dataset}
	%	\vspace{-2.0em}
\end{figure}

\subsubsection{Effect of Dataset with Augmented Features}
To validate the effectiveness of the proposed COVID-19 dataset with augmented features, we train the best-generated model (M4) with the new dataset and compare it with the original one. The learning curve plot of the two settings is illustrated in Fig.\ref{fig:Epoch_Loss_Two_Dataset}.
As it is evident in Fig.\ref{fig:Epoch_Loss_Two_Dataset}, the validation loss of the single feature data doesn't improve much after the 1000th epoch. However, it can be observed that in the train and validation loss obtained by learning the new dataset, the model almost keeps the regular decreasing trend till the last epoch. Also, the final validation loss for the model trained with the new data is superior over the model trained by the original data. The training and validation loss of the original and new datasets are provided in Table .\ref{tab:NewDataOldDataTrainValidation}.
Another important thing to interpret from Fig.\ref{fig:Epoch_Loss_Two_Dataset} is the overfitting of the best model on the original data in a short time after about the 400th epoch. This also shows that the model wasn't capable of finding sufficient distinguished features on the training dataset to increase the accuracy of forecasting cases on validation data. 

\begin{table}[h]
	%	\vspace{-1.5em}
	\renewcommand{\arraystretch}{1.1}
	%\small
	\tabcolsep=0.08cm
	\begin{center}
		\caption{Mean train and validation RMSE loss obtained by 3 individual runs over the new dataset with augmented features and the original data.}\label{tab:NewDataOldDataTrainValidation}
		\vspace{0em}
		\scalebox{1} {
			\begin{tabular}{lcc} % m{1.64cm}m{1.64cm}m{1.64cm}m{1.64cm}m{1.64cm}
				\hline
				Model & \hspace{1.1em}Train Loss & \hspace{1.1em}Validation Loss \\ 
				
				%	\multicolumn{1}{c}{index} & \multicolumn{1}{c}{cases} & \multicolumn{1}{c}{Cumul\_num} & \multicolumn{1}{c}{d\_Status} & \multicolumn{1}{c}{g\_possibility}\\
				\hline
				
				M4+Our Dataset &\hspace{1.1em} \textbf{1.34e-3} &\hspace{1.1em} \textbf{1.64e-3}   \\
				M4+Original Dataset &\hspace{1.1em} 3.05e-3 &\hspace{1.1em} 3.39e-3   \\
				\hline
			\end{tabular}
		}
	\end{center}
	\vspace{-0.0em}
\end{table}

\section{Conclusions}\label{conclusionssection}

In this paper, we proposed a new approach and dataset to forecast the daily cases of COVID-19 more accurately than the common approaches that are based on trial and error in finding an acceptable architecture. We also mentioned the limitations caused by the data-hungry problem in deep learning models and investigated how the proposed approach guarantee to find an architecture that can provide a promising solution despite having limited data such as COVID-19 cases. To validate our proposed approach, NAS-BBA, we provided a set of detailed experiments and compared the results of different custom architectures and the ones obtained by the NAS-BBA framework. In all the cases, the results validate the proposed approach's effectiveness in finding the best deep architecture for forecasting COVID-19 cases. Finally, we trained the best-generated model on the original data and the one proposed in this paper. The results indicate that our proposed dataset with augmented features provides a significant improvement to the model. 

From this paper, there can be several topics for the research community that deserve further study. First, as we also mentioned in the experiment section, the deep model training phase for BBA individual fitness evaluation is time-consuming. One can find utilizing an alternative method for the training phase that is more accurate and faster. Secondly, the impact of other hyperparameters such as learning rates, regularization lambda coefficient, or optimization method on improving the final model can be studied. Also, an alternate optimization method can be utilized \cite{Rahbar2020} for tuning the hyperparameters. Lastly, in this paper, we employed vanilla LSTM units to forecast COVID-19 cases. In future research, other variants of LSTM, recurrent units, and advanced structures such as the combination of convolution and recurrent neural networks can be utilized, and their efficacy can be studied.

\bibliographystyle{unsrtnat}
\bibliography{paper}  %%% Uncomment this line and comment out the ``thebibliography'' section below to use the external .bib file (using bibtex) .

\begin{thebibliography}{10}

\bibitem{Pathan2020}
Refat~Khan Pathan, Munmun Biswas, and Mayeen~Uddin Khandaker.
\newblock Time series prediction of covid-19 by mutation rate analysis using
  recurrent neural network-based lstm model.
\newblock {\em Chaos, solitons, and fractals}, 138:110018, Sep 2020.

\bibitem{Arora2020}
Parul Arora, Himanshu Kumar, and Bijaya~Ketan Panigrahi.
\newblock Prediction and analysis of covid-19 positive cases using deep
  learning models a descriptive case study of india.
\newblock {\em Chaos, solitons, and fractals}, 139:110017, Oct 2020.

\bibitem{Lee2020}
Junghwan Lee, Casey Ta, Jae~Hyun Kim, Cong Liu, and Chunhua Weng.
\newblock Severity prediction for covid-19 patients via recurrent neural
  networks.
\newblock {\em medRxiv}, 2020.

\bibitem{Hawas2020}
Mohamed Hawas.
\newblock Generated time-series prediction data of covid-19s daily infections
  in brazil by using recurrent neural networks.
\newblock {\em Data in Brief}, 32:106175, 2020.

\bibitem{Lalmuanawma2020}
Samuel Lalmuanawma, Jamal Hussain, and Lalrinfela Chhakchhuak.
\newblock Applications of machine learning and artificial intelligence for
  covid-19 (sars-cov-2) pandemic: A review.
\newblock {\em Chaos, Solitons and Fractals}, 139:110059, 2020.

\bibitem{Ke2020}
Yi-Yu Ke, Tzu-Ting Peng, Teng-Kuang Yeh, Wen-Zheng Huang, Shao-En Chang,
  Szu-Huei Wu, Hui-Chen Hung, Tsu-An Hsu, Shiow-Ju Lee, Jeng-Shin Song,
  Wen-Hsing Lin, Tung-Jung Chiang, Jiunn-Horng Lin, Huey-Kang Sytwu, and
  Chiung-Tong Chen.
\newblock Artificial intelligence approach fighting covid-19 with repurposing
  drugs.
\newblock {\em Biomedical Journal}, 2020.

\bibitem{Tuli2020}
Shreshth Tuli, Shikhar Tuli, Rakesh Tuli, and Sukhpal~Singh Gill.
\newblock Predicting the growth and trend of covid-19 pandemic using machine
  learning and cloud computing.
\newblock {\em Internet of Things}, 11:100222, 2020.

\bibitem{ArunKumar2021}
K.E. ArunKumar, Dinesh~V. Kalaga, Ch. Mohan~Sai Kumar, Masahiro Kawaji, and
  Timothy~M Brenza.
\newblock Forecasting of covid-19 using deep layer recurrent neural networks
  (rnns) with gated recurrent units (grus) and long short-term memory (lstm)
  cells.
\newblock {\em Chaos, Solitons and Fractals}, 146:110861, 2021.

\bibitem{Gautam2021}
Yogesh Gautam.
\newblock Transfer learning for covid-19 cases and deaths forecast using lstm
  network.
\newblock {\em ISA Transactions}, 2021.

\bibitem{Shastri2021}
Sourabh Shastri, Kuljeet Singh, Sachin Kumar, Paramjit Kour, and Vibhakar
  Mansotra.
\newblock Deep-lstm ensemble framework to forecast covid-19: an insight to the
  global pandemic.
\newblock {\em International Journal of Information Technology}, 2021.

\bibitem{Abbasimehr2021}
Hossein Abbasimehr and Reza Paki.
\newblock Prediction of covid-19 confirmed cases combining deep learning
  methods and bayesian optimization.
\newblock {\em Chaos, Solitons and Fractals}, 142:110511, 2021.

\bibitem{Chandra2021}
Rohitash Chandra, Ayush Jain, and Divyanshu~Singh Chauhan.
\newblock Deep learning via {LSTM} models for {COVID-19} infection forecasting
  in india.
\newblock {\em CoRR}, abs/2101.11881, 2021.

\bibitem{Salgotra2020}
Rohit Salgotra, Mostafa Gandomi, and Amir~H. Gandomi.
\newblock Evolutionary modelling of the covid-19 pandemic in fifteen most
  affected countries.
\newblock {\em Chaos, Solitons and Fractals}, 140:110118, 2020.

\bibitem{Salgotra2020a}
Rohit Salgotra, Mostafa Gandomi, and Amir~H. Gandomi.
\newblock Time series analysis and forecast of the covid-19 pandemic in india
  using genetic programming.
\newblock {\em Chaos, solitons, and fractals}, 138:109945, Sep 2020.

\bibitem{Chimmula2020}
Vinay Kumar~Reddy Chimmula and Lei Zhang.
\newblock Time series forecasting of covid-19 transmission in canada using lstm
  networks.
\newblock {\em Chaos, Solitons and Fractals}, 135:109864, 2020.

\bibitem{daSilva2020}
Ramon~Gomes {da Silva}, Matheus Henrique Dal~Molin Ribeiro, Viviana~Cocco
  Mariani, and Leandro dos Santos~Coelho.
\newblock Forecasting brazilian and american covid-19 cases based on artificial
  intelligence coupled with climatic exogenous variables.
\newblock {\em Chaos, Solitons and Fractals}, 139:110027, 2020.

\bibitem{Yang2010}
Xin-She Yang.
\newblock {\em A New Metaheuristic Bat-Inspired Algorithm}, pages 65--74.
\newblock Springer Berlin Heidelberg, Berlin, Heidelberg, 2010.

\bibitem{Mirjalili2014}
Seyedali Mirjalili, Seyed~Mohammad Mirjalili, and Xin-She Yang.
\newblock Binary bat algorithm.
\newblock {\em Neural Computing and Applications}, 25(3):663--681, 2014.

\bibitem{Xiong2018}
Zhaohan Xiong, Martyn~P Nash, Elizabeth Cheng, Vadim~V Fedorov, Martin~K
  Stiles, and Jichao Zhao.
\newblock {ECG} signal classification for the detection of cardiac arrhythmias
  using a convolutional recurrent neural network.
\newblock {\em Physiological Measurement}, 39(9):094006, sep 2018.

\bibitem{Rather2015}
Akhter~Mohiuddin Rather, Arun Agarwal, and V.N. Sastry.
\newblock Recurrent neural network and a hybrid model for prediction of stock
  returns.
\newblock {\em Expert Systems with Applications}, 42(6):3234 -- 3241, 2015.

\bibitem{Liu2015}
Shujie Liu, Shujie Liu, Nan Yang, Mu~Li, and Ming Zhou.
\newblock A recursive recurrent neural network for statistical machine
  translation.
\newblock In {\em A Recursive Recurrent Neural Network for Statistical Machine
  Translation}. ACL, June 2015.

\bibitem{Chung2014}
Junyoung Chung, {\c{C}}aglar G{\"{u}}l{\c{c}}ehre, KyungHyun Cho, and Yoshua
  Bengio.
\newblock Empirical evaluation of gated recurrent neural networks on sequence
  modeling.
\newblock {\em CoRR}, abs/1412.3555, 2014.

\bibitem{Schuster1997}
M.~{Schuster} and K.~K. {Paliwal}.
\newblock Bidirectional recurrent neural networks.
\newblock {\em IEEE Transactions on Signal Processing}, 45(11):2673--2681, Nov
  1997.

\bibitem{Soltani2016}
Rohollah Soltani and Hui Jiang.
\newblock Higher order recurrent neural networks.
\newblock {\em CoRR}, abs/1605.00064, 2016.

\bibitem{ShivaPrakash2019}
B.~Shiva~Prakash, K.~V. Sanjeev, Ramesh Prakash, and K.~Chandrasekaran.
\newblock A survey on recurrent neural network architectures for sequential
  learning.
\newblock In Jagdish~Chand Bansal, Kedar~Nath Das, Atulya Nagar, Kusum Deep,
  and Akshay~Kumar Ojha, editors, {\em Soft Computing for Problem Solving},
  pages 57--66, Singapore, 2019. Springer Singapore.

\bibitem{Hochreiter1997}
Sepp Hochreiter and J\"{u}rgen Schmidhuber.
\newblock Long short-term memory.
\newblock {\em Neural Comput.}, 9(8):1735–1780, November 1997.

\bibitem{Cirrincione2020}
Cirrincione L, Plescia F, Ledda C, Rapisarda V, Martorana D, Moldovan R.E,
  Theodoridou K, and Cannizzaro.
\newblock Covid-19 pandemic prevention and protection measures to be adopted at
  the workplace.
\newblock {\em Sustainability}, 2020.

\bibitem{Tay2020}
Keng Jin Darren~Tay et~al.
\newblock Trauma and orthopaedics in the covid-19 pandemic breaking every wave.
\newblock {\em Singapore Medical Journal}, 2020.

\bibitem{Koh2020}
David Koh.
\newblock Covid-19 lockdowns throughout the world.
\newblock {\em Occupational Medicine}, 70(5):322--322, 05 2020.

\bibitem{Gupta2019}
Deepak Gupta, Jatin Arora, Utkarsh Agrawal, Ashish Khanna, and Victor Hugo~C.
  {de Albuquerque}.
\newblock Optimized binary bat algorithm for classification of white blood
  cells.
\newblock {\em Measurement}, 143:180 -- 190, 2019.

\bibitem{Nakamura2012}
R.~Y.~M. {Nakamura}, L.~A.~M. {Pereira}, K.~A. {Costa}, D.~{Rodrigues}, J.~P.
  {Papa}, and X.~. {Yang}.
\newblock Bba: A binary bat algorithm for feature selection.
\newblock In {\em 2012 25th SIBGRAPI Conference on Graphics, Patterns and
  Images}, pages 291--297, Aug 2012.

\bibitem{Zoph2016}
Barret Zoph and Quoc~V. Le.
\newblock Neural architecture search with reinforcement learning.
\newblock {\em CoRR}, abs/1611.01578, 2016.

\bibitem{Stanley2002}
Kenneth~O. Stanley and Risto Miikkulainen.
\newblock Evolving neural networks through augmenting topologies.
\newblock {\em Evolutionary Computation}, 10(2):99--127, 2002.

\bibitem{Rahbar2020}
Mahdi Rahbar and Samaneh Yazdani.
\newblock Historical knowledge-based mbo for global optimization problems and
  its application to clustering optimization.
\newblock {\em Soft Computing}, 2020.

\end{thebibliography}





% \end{thebibliography}

\end{document}